\pdfoutput=1
\documentclass[11pt]{article}

\newif\ifarxiv
\arxivtrue % make it 'true' or 'false' if you want to keep the anonymized or public version

\ifarxiv
    \usepackage{acl}
\else
    \usepackage[review]{acl}
\fi

\usepackage{times}
\usepackage{latexsym}
\usepackage{url}
\usepackage[T1]{fontenc}
\usepackage[utf8]{inputenc}

\usepackage{microtype}

\usepackage{inconsolata}

\usepackage{graphicx}

\usepackage{enumitem}
\usepackage{array}
\usepackage{amsthm}
\usepackage{booktabs} % For nice rules in tables

\usepackage{listings}
\usepackage{longtable}
\usepackage{booktabs}
\usepackage{soul,xcolor}
\usepackage{longtable}
\sethlcolor{yellow}
\usepackage{subcaption}
\usepackage{multirow}
\usepackage{bm}
\usepackage{svg}

\usepackage{calligra}

\usepackage{tikz}
\newcommand{\grayb}{\textcolor{gray}{50.00}}
\newcommand{\grayz}{\textcolor{gray}{0.00}}
\definecolor{taborange}{rgb}{1.0, 0.498, 0.0549}
\definecolor{tabblue}{rgb}{0.1216, 0.4667, 0.7059}

 \usepackage[most]{tcolorbox}

\newtcolorbox{myquotebox}{
  colback=white!0, % Transparent background
  colframe=black, % Black frame
  rounded corners,
  boxrule=0.5pt, % Frame thickness
  title=Prompt:,
  left=2mm, % Left margin within the box
  right=2mm, % Right margin within the box
  top=1mm, % Top margin within the box
  bottom=1mm % Bottom margin within the box
}

\usepackage{todonotes}
\usepackage{xspace}
\definecolor{lightgrey}{RGB}{158, 158, 158}
\definecolor{goldenrod}{rgb}{0,0,0.8}
\definecolor{deepred}{rgb}{0.6,0,0}
\definecolor{deepgreen}{rgb}{0,0.5,0}
\definecolor{pink}{RGB}{219, 48, 122}
\definecolor{forestgreen}{RGB}{34,139,34}
\definecolor{goldenrod}{RGB}{218,165,32}
\definecolor{sepia}{RGB}{112,66,20}

\usepackage[capitalise]{cleveref}
\crefname{figure}{Fig.}{Figs.}
\crefname{table}{Table}{Tables}
\crefname{appendix}{App.}{App.}
\crefname{section}{§}{§§}
\crefname{equation}{Eq.}{Eqs.}

\newcommand{\algname}{$\text{UNIT}_\text{ref}$}
\newcommand{\algnameCut}{$\text{UNIT}_\text{cut}$}

\newcommand{\upgreen}{\textcolor{green!50!black}{$\uparrow$}}
\newcommand{\downred}{\textcolor{red}{$\downarrow$}}
\newcommand{\stepone}{\raisebox{-0.4ex}{\includegraphics[width=0.9em]{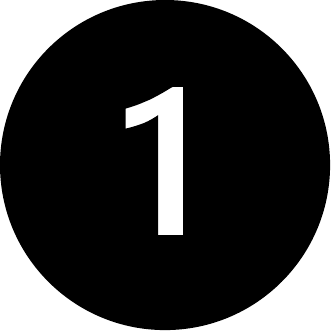}}}
\newcommand{\steptwo}{\raisebox{-0.4ex}{\includegraphics[width=0.9em]{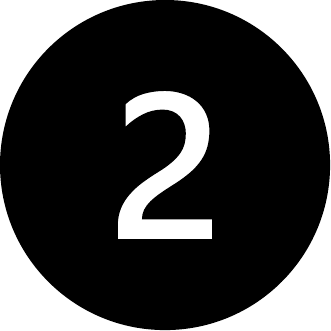}}}
\newcommand{\stepthree}{\raisebox{-0.4ex}{\includegraphics[width=0.9em]{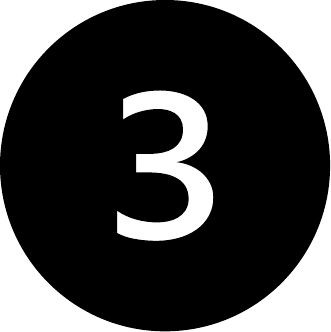}}}
\newcommand{\stepfour}{\raisebox{-0.4ex}{\includegraphics[width=0.9em]{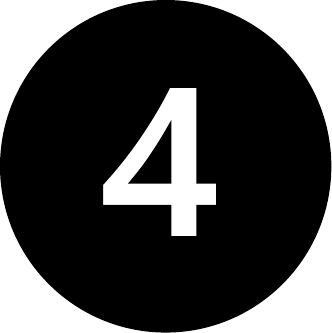}}}

\definecolor{lightred}{RGB}{254, 138, 138}
\definecolor{lightblue}{RGB}{176, 195, 248}
\definecolor{lightgreen}{RGB}{138, 218, 174}
\newcommand\myparagraph[1]{
\vskip 0.05in 
\noindent{\bf {#1}}}

\definecolor{boxborder}{RGB}{86, 113, 209}  %  border
\definecolor{boxbg}{RGB}{255, 255, 255}    % background
\definecolor{boxtitle}{RGB}{255, 255, 255} %  title text
\definecolor{boxheader}{RGB}{86, 113, 209}  %  header

\newcommand*{\crosssymbol}{%
    \text{%
      \raise 1ex\hbox{%
        \rlap{\vrule height.2pt depth.2pt width .75ex}%
        \hbox to .75ex{\hss\vrule height .5ex depth 1ex\hss}%
      }%
    }%  
}
\usepackage{enumitem}
\usepackage{makecell}
\setlist[itemize]{leftmargin=5pt}

\title{Balancing Truthfulness and Informativeness with Uncertainty-Aware Instruction Fine-Tuning}

\author{
 \textbf{Tianyi Wu\textsuperscript{$N$ $Z$}\thanks{Equal contributions.}}\quad
 \textbf{Jingwei Ni\textsuperscript{$E$ $Z$}\footnotemark[1]}\quad
 \textbf{Bryan Hooi\textsuperscript{$N$}\thanks{Equal co-supervision in a dice-rolled order.}}\quad
 \textbf{Jiaheng Zhang\textsuperscript{$N$}\footnotemark[2]}\quad
\\
 \textbf{Elliot Ash\textsuperscript{$E$}\footnotemark[2]}\quad
 \textbf{See-Kiong Ng\textsuperscript{$N$}\footnotemark[2]}\quad
 \textbf{Mrinmaya Sachan\textsuperscript{$E$\footnotemark[2]}}\quad
 \textbf{Markus Leippold\textsuperscript{$Z$ $S$\footnotemark[2]}}
\\
 \textsuperscript{$E$}ETH Zürich\quad
 \textsuperscript{$N$}National University of Singapore\quad \\
 \textsuperscript{$Z$}University of Zürich\quad
 \textsuperscript{$S$}Swiss Finance Institute (SFI)
\\
 \small{
   \href{tianyi_wu@u.nus.edu}{tianyi\_wu@u.nus.edu}\quad \href{jingni@ethz.ch}{jingni@ethz.ch}\quad \href{markus.leippold@df.uzh.ch}{markus.leippold@df.uzh.ch}
 }
}

\begin{document}
\maketitle

\begin{abstract}
Instruction fine-tuning (IFT) can increase the informativeness of large language models (LLMs), but may reduce their truthfulness. This trade-off arises because IFT steers LLMs to generate responses containing long-tail outputs that include knowledge that may not be well-covered during pre-training. As a result, models become more informative but less accurate when generalizing to unseen tasks. In this paper, we empirically demonstrate how unfamiliar knowledge in IFT datasets can undermine the truthfulness of LLMs, and propose two new IFT paradigms, \algnameCut\space and \algname, to address this issue. \algnameCut\ detects and removes unfamiliar knowledge from IFT data to enhance truthfulness, while \algname\ adds a reflection section that flags uncertain claims. Experiments demonstrate that \algnameCut\ substantially improves truthfulness and \algname\ preserves informativeness while addressing hallucinations by signaling uncertainty. We open-source all relevant resources to facilitate future research at \url{https://github.com/AndrewWTY/UNIT}.
% \algnameCut\space identifies and removes unfamiliar knowledge from IFT datasets to mitigate its impact on model truthfulness, whereas \algname\space trains LLMs to recognize their uncertainty and explicitly indicate it at the end of their responses. Our experiments show that \algnameCut\space substantially improves LLM truthfulness, while \algname\space maintains high informativeness and reduces hallucinations by distinguishing between confident and uncertain statements.\footnote{We will open-source all data, code, and training recipes.}
\end{abstract}

% paper rewrite:
% From openrview:
% Qwen 2.5 full set experiment, explanation
% Validation check of GPT output
% The choice of MT-Bench to evaluate informativeness
% Explain why balance needs to be achieved during the SFT stage in introduction
% Marginal Improvement, show maximum possible improvement and explain.
% Explain the choice of training datasets.

% Others:
% Related work (?)
% Examples Illustration of UNIT

\section{Introduction}

General-purpose alignment pursues responses that ``provide a clear, complete, and detailed answer with additional information valuable for users.'' \citep{zheng2023judgingllmasajudgemtbenchchatbot}, where informativeness plays a critical role. 
To encourage detailed, user-valuable responses, prior works have invested significant effort in collecting high-quality Instruction Fine-Tuning (IFT) data and fine-tuning LLMs on them \citep{zhao2024longalignmentsimpletoughtobeat,liu2024makesgooddataalignment,zhou2023limaalignment}.
However, using such high-quality IFT data to steer LLMs to be informative might harm their truthfulness, as they might be taught to extrapolate beyond their parametric knowledge to provide extensive details. % that may not be familiar to their parametric knowledge. 
For example, in LIMA \citep{zhou2023limaalignment}, LLMs are tuned to cite ``(Klämbt, 2009)'' to support the claim that ``brain glial cells migrate.''---a reference likely too niche to be familiarised as part of the model's parametric knowledge during pre-training. This knowledge gap between pre-training and fine-tuning may encourage LLMs to generate informative, confident, but inaccurate answers when generalizing to unseen tasks, inducing confident hallucinations \citep{gekhman-etal-2024-fine,kang2024unfamiliarfinetuningexamplescontrol}. 
\begin{figure}[t]
    \centering
    \includegraphics[width=0.43\textwidth]{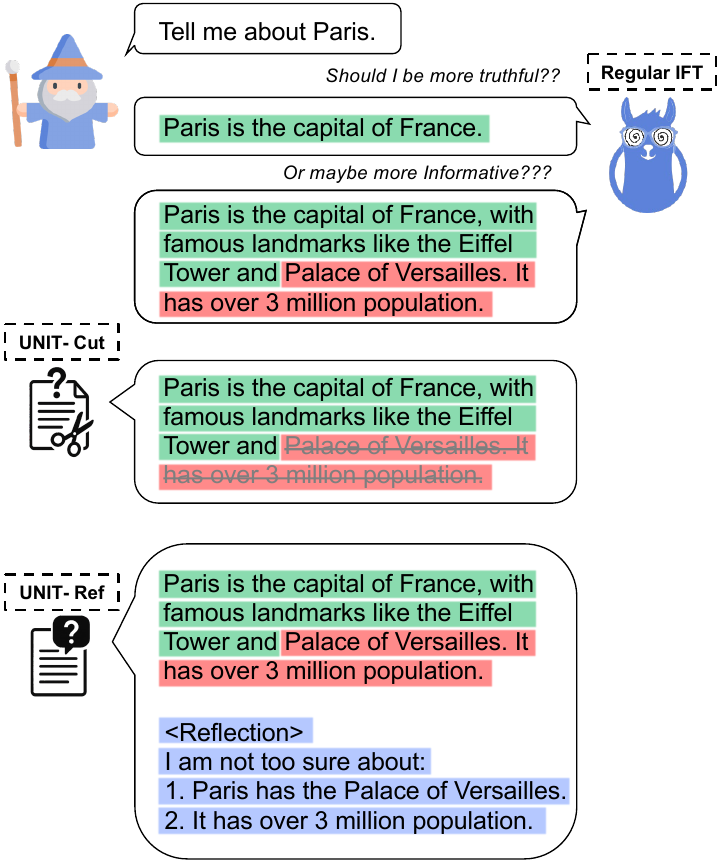}
    \caption{Right: in regular IFT, tuning LLMs for better informativeness may encourage LLMs to produce \colorbox{lightred}{uncertain claims} that are less likely to be correct than \colorbox{lightgreen}{certain claims}. Left: In uncertain-aware IFT, LLMs are tuned to either leave out uncertain claims (\algnameCut) or \colorbox{lightblue}{reflect on them} (\algname) while maintaining informativeness.}
    \label{fig:overview}
    \vspace{-1em}
\end{figure}

To mitigate hallucinations, previous work focuses on constructing specialized, honesty-oriented training data to improve LLM truthfulness, achieving promising results \citep{zhang2024rtuninginstructinglargelanguage, yang2024alignmenthonesty, band2024linguisticcalibrationlongformgenerations}. However, these methods largely overlook how to safely fine-tune LLMs on existing high-quality IFT data to enhance informativeness without reducing truthfulness. This gap is particularly relevant for practitioners who depend on IFT to adapt LLMs to specific downstream tasks or domains \citep{niklaus2025lawinstructresourcestudyinglanguage,wu2024susgengptdatacentricllmfinancial}.

In this work, we uncover how using high-quality instruction fine-tuning data affects the truthfulness of LLMs, and how to safely use such data without compromising truthfulness. Particularly, we investigate this by answering two research questions in a logical order:

\myparagraph{RQ1. Does unfamiliar knowledge in human-annotated IFT datasets affect truthfulness?} IFT achieves generalizable informativeness by having detailed long-form generation with diverse, factually-accurate, information-dense instruction-response pairs (e.g., LIMA). However, it remains unclear whether fine-tuning on this dense information that may be unfamiliar to LLMs' parametric knowledge would cost a trade-off in the truthfulness of LLMs \citep{zhao2024longalignmentsimpletoughtobeat}.

We first propose (\underline{UN}certainty-aware \underline{I}nstruction \underline{T}uning by \underline{Cut}ting) \algnameCut, where we pinpoint the LLMs' uncertainty about the claims in an instruction fine-tuning dataset and reconstruct a ``more familiar'' variant of the IFT dataset for instruction fine-tuning. Our findings reveal that incorporating more unfamiliar knowledge in IFT reduces model truthfulness. Furthermore, by removing uncertain knowledge within the IFT dataset using \algnameCut, the truthfulness of the responses increases, while the informativeness might be negatively affected.

\myparagraph{RQ2. How can we leverage original high-quality IFT data without compromising trustworthiness?}
While \algnameCut\space improves truthfulness by brutally removing unfamiliar knowledge from IFT data, this may hurt informativeness by sacrificing valuable details or structures. This raises a key question: can we directly apply the informative, elaborately crafted IFT datasets from prior work while mitigating hallucinations?
To address this, we propose the second algorithm \algname\ (\underline{UN}certainty-aware \underline{I}nstruction \underline{T}uning by \underline{Ref}lecting), an IFT paradigm that fine-tunes models to report their uncertainty after responses. Specifically, instead of reconstructing a less uncertain variant dataset like \algnameCut, \algname\space appends a ``reflection'' after each response that lists the uncertain claims to teach the model to reflect on its uncertainty. This approach preserves the original IFT datasets' response information richness while promoting honesty: the model learns to signal uncertainty, enabling users to perform targeted post-verification of specific claims. A comparison between regular IFT and \algname\space can be found in \cref{fig:overview}. 

In summary, our contributions are: (1) We empirically demonstrate that fine-tuning on unfamiliar knowledge in high-quality IFT datasets always reduces the truthfulness of LLMs' responses, and find removing unfamiliar knowledge (\algnameCut) an effective algorithm to reduce IFT-caused hallucination (\cref{sec:RQ1}). (2) We introduce \algname, an IFT paradigm that takes advantage of the original high-quality IFT datasets to improve informativeness while preserving trustworthiness by flagging uncertain claims (\cref{sec:RQ2}).
\section{Related Work}
\begin{figure*}[t]
    \centering	\includegraphics[width=0.95\textwidth]{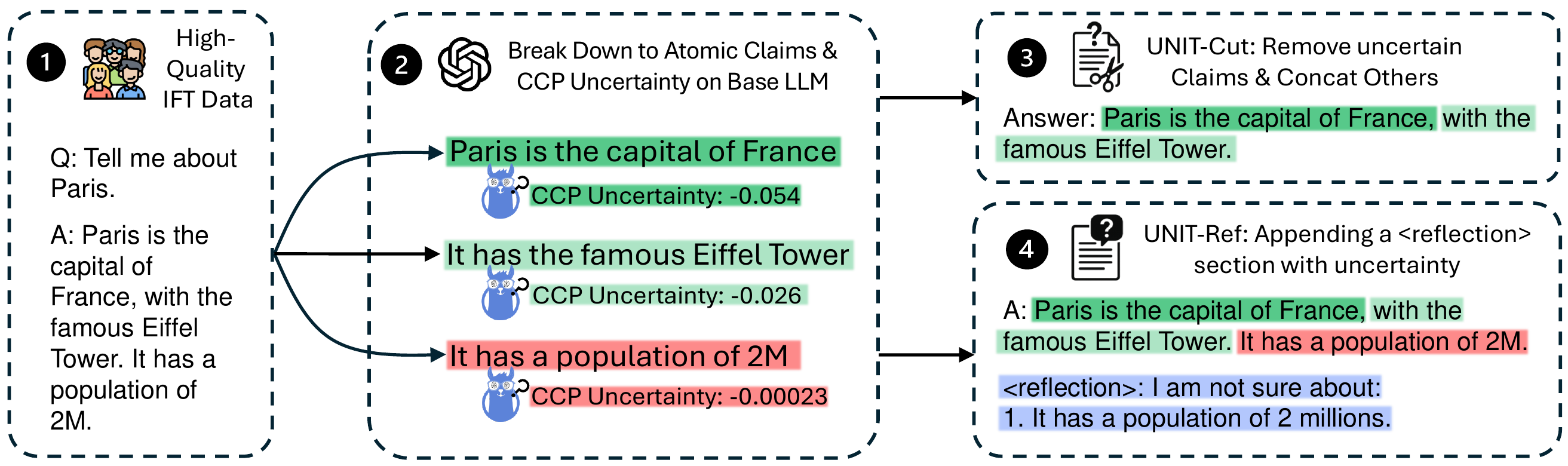}
    \vspace{-2mm}
    \caption{Illustration of the procedure of \algnameCut\space and \algname. Given a high-quality IFT dataset (e.g., manually annotated like LIMA and LFRQA), we first measure claim-level uncertainty for each response (\steptwo). Then we construct new IFT data by removing the uncertain claims (\stepthree\ \algnameCut) or reflecting them after the responses (\stepfour\ \algname).}
	\label{fig:illustration}
\vspace{-1em}
\end{figure*}

\myparagraph{LLM Honesty Alignment.} Various previous works investigate how to align LLMs to appropriately express their knowns and unknowns \citep{yang2024logulongformgenerationuncertainty,yang2024alignmenthonesty,xu2024rejectionimprovesreliabilitytraining,cheng2024aiassistantsknowdont,zhang2024rtuninginstructinglargelanguage,band2024linguisticcalibrationlongformgenerations}. Typically this involves (1) prompting the model with information‐seeking queries; (2) grading its responses against the ground truth; and (3) fine‐tuning the model to express higher confidence for correct answers and lower confidence for incorrect ones. Confidence indicators include refusals \citep{zhang2024rtuninginstructinglargelanguage}, numerical confidence scores \citep{yang2024alignmenthonesty}, or linguistic uncertainty markers \citep{band2024linguisticcalibrationlongformgenerations,yang2024logulongformgenerationuncertainty}. While these works focus on constructing their own honesty-oriented training data, they largely overlook how to safely leverage high-quality IFT data \citep{zhou2023limaalignment} to improve other aspects of LLM performance---such as informativeness---without compromising truthfulness. This challenge is particularly relevant for practitioners who rely on elaborate IFT data to adapt LLMs to specific tasks or domains \citep{niklaus2025lawinstructresourcestudyinglanguage,wu2024susgengptdatacentricllmfinancial,fatemi2024comparativeanalysisinstructionfinetuning,zhang2024alpacareinstructiontunedlargelanguagemodels}. Addressing this gap, our work first demonstrates that unfamiliar knowledge in high-quality IFT data can induce hallucination, and we propose \algnameCut\ and \algname\space as a safer paradigm for leveraging such data. Our work takes the first step to improve LLM honesty while considering other dimensions, taking informativeness as a case study.

\myparagraph{IFT's Impact on Informativeness and Truthfulness.} Prior work in IFT alignment emphasizes informativeness as a core element for helpfulness. For example, \citet{zhou2023limaalignment} collects ``complete and detailed'' responses; \citet{liu2024makesgooddataalignment} enhances response helpfulness by ``depth and details''; and \citet{Zhao2024LongIM} finds that longer responses can benefit alignment. Hence, our work focuses on how to benefit from informative IFT data without compromising truthfulness. \citet{gekhman-etal-2024-fine} find that IFT rarely increases hallucination with early stop (e.g., under 5 epochs). \citet{zhao2024longalignmentsimpletoughtobeat} report that IFT does not degrade performance on factual-knowledge benchmarks. However, we find that even with small epoch numbers, incorporating unfamiliar knowledge can still harm truthfulness.
\myparagraph{Uncertainty Measurement for LLMs.} Uncertainty measurement is a critical technique for detecting hallucinations, since higher uncertainty often corresponds to lower generation quality \citep{xiong2024llmsexpressuncertaintyempirical,vashurin2025benchmarkinguncertaintyquantificationmethods}. Uncertainty measures can be divided into two types: sequence-level measures, which evaluate uncertainty across entire generated sequences \citep{kuhn2023semanticuncertaintylinguisticinvariances,duan-etal-2024-shifting}, and claim-level measures, which assess uncertainty for individual factual claims \citep{fadeeva-etal-2024-fact}. In this work, we adopt Claim-Conditioned Probability (CCP) \citep{fadeeva-etal-2024-fact}, identified by \citet{vashurin2025benchmarkinguncertaintyquantificationmethods} as the best-performing claim-level measure, to quantify LLMs' familiarity with the claims in high-quality IFT data.

\section{RQ1: Does Unfamiliar Knowledge in IFT Affect Truthfulness?} \label{sec:RQ1}

To investigate RQ1, we design controlled experiments comparing IFT outcomes before and after removing unfamiliar knowledge. We first introduce \algnameCut\ (\cref{sec:unit_cut}), an algorithm that fine-tunes LLMs on IFT data with unfamiliar content removed. We then introduce the training datasets (\cref{sec:training_data}) and experimental settings (\cref{sec:rq1_ex_details}). Finally, we present the results and key takeaways (\cref{sec:section2_result}).

\subsection{Methodology - \algnameCut} \label{sec:unit_cut}
In this section, we introduce \algnameCut\ (illustrated in \cref{fig:illustration} \steptwo\ and \stepthree), an IFT paradigm with unfamiliar knowledge removed. The training data construction of \algnameCut\space consists of three steps: First, it measures CCP-based uncertainty \citep{fadeeva-etal-2024-fact} for atomic claims in IFT data responses. Second, it categorizes claims into familiar or unfamiliar based on a given uncertainty threshold. Third, it concatenates the familiar claims into a new response using a language model rewriter, and fine-tunes the target LLM. The procedure is detailed as follows. 

 \paragraph{Firstly: Finding Unfamiliar Atomic Claims with CCP.} 
Given an instruction dataset $D$ and a target LLM $M$, our first step is to measure the uncertainty of all atomic claims in $D$ using the CCP algorithm. The dataset $D$ contains \(N\) instruction-response pairs \(D = \{ (I_i, R_i) \}_{i=1}^{N}\). Let $x_{i,j}$ denotes the $j$-th token in response 
From each response \(R_i\), CCP extracts a set of atomic factual claims \(\mathcal{C}_i = \{ C_{i,1}, C_{i,2}, \ldots, C_{i,m_i} \}\), where each \(C_{i,j} \subset R_i\) representing a coherent factual statement. 
For each token \(x_{i,j}\) in a claim \(C_{i,j}\), the target model \(M\) samples the top-\(K\) alternatives
\[
\mathcal{A}(x_{i,j}) = \{ x_{i,j}^1, x_{i,j}^2, \ldots, x_{i,j}^K \},
\]
with probabilities \(P(x_{i,j}^k \mid x_{i,<j})\), where \(x_{i,<j} = \{ x_{i,1}, \ldots, x_{i,j-1} \}\). A natural language inference (NLI) model then assesses the semantic relationship between each alternative \(x_{i,j}^k\) and the original token \(x_{i,j}\) by comparing the contexts $x_{i,<j} \circ x_{i,j}^k$ and $x_{i,1:j} = x_{i,<j} \circ x_{i,j}$, assigning one of three labels: entailment (\(e\)), contradiction (\(c\)), or neutral (\(n\)). Define:

{\small
\begin{align*}
\texttt{Me}(x_{i,j}) &= \left\{ x_{i,j}^k \;\middle|\; 
\text{NLI}(x_{i,<j} \circ x_{i,j}^k,\, x_{i,1:j}) = e \right\} \\
\texttt{CT}(x_{i,j}) &= \left\{ x_{i,j}^k \;\middle|\; 
\text{NLI}(x_{i,<j} \circ x_{i,j}^k,\, x_{i,1:j}) \in \{e, c\} \right\}
\end{align*}
}
where $\texttt{Me}$ (Meaning) intuitively denotes alternative tokens with the same meaning as $x_{i,j}$, while $\texttt{CT}$ (ClaimType) denotes alternative tokens with the same claim type but may contradict $x_{i,j}$. The token-level uncertainty is computed as
\[
\text{CCP}(x_{i,j}) = \frac{\sum_{x_{i,j}^k \in Me(x_{i,j})} P(x_{i,j}^k \mid x_{i,<j})}{\sum_{x_{i,j}^l \in CT(x_{i,j})} P(x_{i,j}^l \mid x_{i,<j})},
\]
and the overall uncertainty for a claim is aggregated by
\[
\text{CCP}_{\text{claim}}(C_{i,j}) = 1 - \prod_{x \in C_{i,j}} \text{CCP}(x).
\]

Using CCP, for each response \(R_i\) we obtain a set of atomic claims with their corresponding uncertainty values, i.e.,
\(
\{ (C_{i,j}, \text{CCP}_{\text{claim}}(C_{i,j})) \}_{j=1}^{m_i}
\)
, where a higher CCP value means the model is more uncertain about each claim.

\paragraph{Secondly: Labeling Uncertain Atomic Claims in Responses.}
In the previous step, we compute CCP-based uncertainty scores $\text{CCP}_{\text{claim}}(C_{i,j})$ for all atomic claims $\mathcal{C}_i = \{ C_{i,j} \}_{j=1}^{m_i}$ extracted from response \(R_i\). The next step is to categorize them as familiar or unfamiliar based on the CCP scores.

To do this, we compute the 75th quantile\footnote{We choose 75th quantile for demonstration simplicity. Theoretically, the threshold can be set to other values to control the conservativeness of the algorithm.} 
 of CCP scores across the entire dataset $D$, and use this value as the threshold \(\tau\) for distinguishing unfamiliar (i.e., uncertain) knowledge to the target LLM:
\[
\tau = Q_{0.75}\left(\{ \text{CCP}_{\text{claim}}(C) \mid C \in \mathcal{C} \}\right).
\]
Then, for each claim \(C \in \mathcal{C}\), we assign its uncertainty label as follows:
\[
\ell(C) = \begin{cases}
\text{uncertain}, & \text{if } \text{CCP}_{\text{claim}}(C) > \tau, \\
\text{certain},   & \text{otherwise.}
\end{cases}
\]

\paragraph{Finally: Removing Unfamiliar Knowledge.}
For each instruction–response pair $(I_i,R_i)$ we collect the claims previously labeled as \emph{certain} into a list, keeping their order in the original response:

$$
\mathcal{L}_i=\bigl[C_{i,j}\bigr]_{j:\,\ell(C_{i,j})=\text{certain}}
$$

We pass the list to an auxiliary language-model rewriter $f_{\text{LLM}}$. Conditioned on both the instruction $I_i$ and the list $\mathcal{L}_i$, the model returns a fluent answer that contains only vetted information:

$$
R_i^{\text{cut}}=f_{\text{LLM}}\!\bigl(I_i,\mathcal{L}_i\bigr)
$$

Substituting every $(I_i,R_i)$ with $(I_i,R_i^{\text{cut}})$ yields the IFT data with unfamiliar knowledge (above an uncertainty threshold) removed:

$$
D^{\text{cut}}=\bigl\{(I_i,R_i^{\text{cut}})\bigr\}_{i=1}^{N}
$$

If all claims in a response $R_i$ are labeled as uncertain, $R_i^{\text{cut}}$ should apologize without providing any information. If all claims in response $R_i$ are labeled as certain, $R_i^{\text{cut}} = R_i$. Finally, we fine-tune the target model $M$ on $D^{\text{cut}}$. This completes \algnameCut, ensuring that no claim with high CCP uncertainty influences the model's subsequent instruction-following behaviour. 

Notably, the original CCP is proposed to measure uncertainty for on-policy data (i.e., LLM generations), while we establish a new use case of it to measure LLMs' familiarity to off-policy data (i.e., elaborate IFT data), and prove it effective.

CCP (and other uncertainty measures) was originally designed for information-seeking queries that decompose into independent factual claims. \citet{vashurin2025benchmarkinguncertaintyquantificationmethods} validates CCP's reliability only on information-seeking tasks, but also shows its limitations on non-information-seeking tasks like math and translation. Extending it to other tasks (e.g., reasoning, creative writing, or editing) can be conceptually inaccurate: e.g., in a 2-step reasoning task of first solving $a$ and $b$ given $a=2+2,\space b=a+1$; if a model answers “$a=3$” (wrong) and then “$b=4$” (correct given the earlier mistake), CCP may flag high uncertainty on the first step but low on the second, conditioned on the generated “$a=3$”. We give more examples where CCP may fail in \cref{app:why_info_only}. Therefore, in this work, we focus our use of CCP on information-seeking IFT data, which is directly relevant to our objective of understanding the impact of unfamiliar knowledge on model truthfulness. We detail our prompts for claim extraction, instruction classification, and LLM rewriting in \cref{app:info_seeking_prompt}.

\subsection{Training Data} 
\label{sec:training_data}

We conduct experiments on two datasets: LIMA \citep{zhou2023limaalignment} and LFRQA \citep{han-etal-2024-rag}. LIMA is selected because of its status as a high-quality, human-annotated instruction-following fine-tuning (IFT) dataset. LFRQA also contains long-form human-written instruction-response pairs, but has a stronger focus on information-seeking tasks, where we can apply CCP to measure claim-level uncertainty. Moreover, the queries of LFRQA span across 7 domains (e.g., biomedical, finance, recreation, technology, etc.), which often require domain-specific niche knowledge to answer. Therefore, LFRQA may effectively emulate real-world domain-specific IFT and help us to investigate potential challenges there. 

Since our exploration focuses on IFT data and data-centric approaches to alleviate IFT-induced hallucinations, we investigate various combinations of datasets to draw statistically significant observations. Therefore, we vary LFRQA from 10\% to 100\% of its examples and apply \algnameCut. We then augment each LFRQA subset with LIMA to assess performance when including general-domain helpfulness data.

\subsection{Evaluation and Training Details} \label{sec:rq1_ex_details}
\noindent \textbf{Truthfulness.}
We use FactScore \citep{min-etal-2023-factscore} and WildFactScore \citep{zhao2024wildhallucinationsevaluatinglongformfactuality} to fact-check atomic claims in LLMs' long-form outputs. 
FactScore prompts LLMs to generate 500 biographies (\emph{Bio}), while WildFactScore prompts to introduce 7K entities absent from Wikipedia (\emph{WildHalu}\footnote{We randomly sample 500 entities from WildHalu for budget control.}). 
FactScore decomposes each text into atomic claims and verifies them using a retrieval-augmented LLM agent. 
The final truthfulness score is the percentage of atomic claims verified as true.

\myparagraph{Informativeness.}
We investigate how changes in informativeness dynamically affect truthfulness by evaluating both dimensions on benchmark prompts used in FactScore and WildFactScore. To facilitate informativeness measurement on FactScore and WildFactScore's benchmark prompts, we adapt the MT-Bench \citep{zhou2023limaalignment} pairwise LLM-judge format, which, although originally designed to assess general helpfulness, is well-suited for our needs. Because Bio and WildHalu tasks are inherently information-seeking, making assessing helpfulness a reasonable proxy for informativeness.

Using GPT-4o as the LLM judge, we present a question with two answers and ask which is more informative or if they are tied—while explicitly instructing the model to ignore truthfulness to isolate the informativeness dimension. To reduce position bias, we randomize the order of answers. All evaluations are conducted relative to a LIMA fine-tuned baseline. The final informativeness score is computed as the win rate plus half the tie rate. Truthfulness and informativeness scoring details are provided in \cref{app:eval_metric_details}. We focus on out-of-distribution (OOD) evaluations using Bio and WildHalu, which do not appear in training, aligning with IFT's goal of generalizing to unseen tasks.

\myparagraph{Training and Inference Details.}
All experiments use full fine-tuning on Llama-3.1-8B \citep{grattafiori2024llama3herdmodels} and Qwen2.5-14B \citep{qwen2.5} for 3 epochs, varying only the IFT data. 
We employ TRL for fine-tuning and vLLM for inference. 
Hyperparameters, chat templates, and other technical details are provided in \cref{app:experiment_details}.

\subsection{Experiment Results} \label{sec:section2_result}
\begin{table}[t]
  \centering
  \scriptsize
  \setlength{\tabcolsep}{3pt} % tighten column padding
  \resizebox{\columnwidth}{!}{
\begin{tabular}{c@{\hspace{2mm}} c@{\hspace{2mm}} l @{\hspace{4mm}} c c c c}
      \toprule
      \multirow{2.5}{*}{\shortstack{\textbf{Model} \\\textbf{/}\\ \textbf{Data}}} 
        & \multirow{2.5}{*}{\textbf{LFRQA\%}} 
        & \multirow{2.5}{*}{\textbf{Method}} 
        & \multicolumn{2}{c}{\textbf{Truth.$\uparrow$}} 
        & \multicolumn{2}{c}{\textbf{Info.$\uparrow$}} \\
      \cmidrule(lr){4-5}\cmidrule(lr){6-7}
      & & 
        & \textbf{Bio.} & \textbf{Wild.}         & \textbf{Bio.} & \textbf{Wild.} \\
      \midrule
    \multirow{8}{*}{\shortstack{Llama3.1-8B\\/\\LFRQA}}
& \multirow{2}{*}{10\%}  & Vanilla     & 56.54   & 82.22   & 22.10 & 25.90 \\
&       & \algnameCut & +3.17\upgreen  & +4.91\upgreen  & -6.10\downred & -9.00\downred \\ 
& \multirow{2}{*}{40\%}  & Vanilla     & 53.41   & 79.00   & 15.60 & 18.75 \\
&       & \algnameCut & +12.12\upgreen & +8.68\upgreen  & -6.10\downred & -8.45\downred \\ 
& \multirow{2}{*}{70\%}  & Vanilla     & 49.15   & 75.32   & 16.80 & 19.70 \\
&       & \algnameCut & +20.88\upgreen & +11.16\upgreen & -6.20\downred & -8.50\downred \\ 
& \multirow{2}{*}{100\%} & Vanilla     & 50.15   & 73.77   & 15.60 & 18.80 \\
&       & \algnameCut & +20.39\upgreen & +15.10\upgreen & -5.60\downred & -8.35\downred \\
\midrule

\multirow{8}{*}{\shortstack{Llama3.1-8B\\/\\LFRQA\\+LIMA}}
% & \multirow{2}{*}{0\%}   & Vanilla     & 44.43   & 77.43   & 50.00 & 50.00 \\
% &       & - uncertain & +10.05\upgreen & -0.96\downred  & -19.60\downred & -14.20\downred \\ 
& \multirow{2}{*}{10\%}  & Vanilla     & 50.97   & 79.43   & 34.70 & 33.75 \\
&       & \algnameCut & +10.30\upgreen & +1.94\upgreen  & -11.70\downred & -8.85\downred \\ 
& \multirow{2}{*}{40\%}  & Vanilla     & 47.51   & 73.89   & 29.90 & 31.70 \\
&       & \algnameCut & +15.98\upgreen & +11.32\upgreen & -14.10\downred & -15.40\downred \\ 
& \multirow{2}{*}{70\%}  & Vanilla     & 44.31   & 73.65   & 30.10 & 26.35 \\
&       & \algnameCut & +16.79\upgreen & +13.34\upgreen & -17.50\downred & -12.90\downred \\ 
& \multirow{2}{*}{100\%} & Vanilla     & 46.81   & 73.04   & 26.30 & 27.35 \\
&       & \algnameCut & +17.45\upgreen & +13.07\upgreen & -13.50\downred & -14.45\downred \\

\midrule
\multirow{8}{*}{\shortstack{Qwen2.5-14B\\/\\LFRQA}}
& \multirow{2}{*}{10\%}  & Vanilla     & 41.14           & 79.70           & 46.30         & 42.50 \\
&       & \algnameCut & +14.32\upgreen  & +2.40\upgreen   & +2.70\upgreen & +8.90\upgreen \\ 
& \multirow{2}{*}{40\%}  & Vanilla     & 51.11           & 81.66           & 50.00         & 41.40 \\
&       & \algnameCut & +10.37\upgreen  & +1.68\upgreen   & -10.40\downred & +7.10\upgreen \\ 
& \multirow{2}{*}{70\%}  & Vanilla     & 50.64           & 80.86           & 45.30         & 39.80 \\
&       & \algnameCut & +6.30\upgreen   & +1.12\upgreen   & -8.40\downred  & +7.80\upgreen \\ 
& \multirow{2}{*}{100\%} & Vanilla     & 49.55           & 80.60           & 47.50         & 39.90 \\
&       & \algnameCut & +4.85\upgreen   & +2.45\upgreen   & -7.80\downred  & +5.30\upgreen \\
\midrule

\multirow{8}{*}{\shortstack{Qwen2.5-14B\\/\\LFRQA\\+LIMA}}
% & \multirow{2}{*}{0\%}   & Vanilla     & 32.43           & 69.06           & 50.00         & 50.00 \\
% &       & - uncertain & +5.94\upgreen   & +6.44\upgreen   & -0.70\downred  & +10.70\upgreen \\ 
& \multirow{2}{*}{10\%}  & Vanilla     & 33.48           & 75.05           & 46.60         & 46.20 \\
&       & \algnameCut & +13.09\upgreen  & +3.32\upgreen   & +1.80\upgreen  & +1.90\upgreen \\ 
& \multirow{2}{*}{40\%}  & Vanilla     & 43.64           & 77.26           & 39.50         & 44.30 \\
&       & \algnameCut & +5.95\upgreen   & +4.35\upgreen   & +7.40\upgreen  & +3.40\upgreen \\ 
& \multirow{2}{*}{70\%}  & Vanilla     & 40.86           & 73.43           & 42.00         & 40.80 \\
&       & \algnameCut & +3.87\upgreen   & +4.06\upgreen   & -3.60\downred  & +4.20\upgreen \\ 
& \multirow{2}{*}{100\%} & Vanilla     & 44.58           & 79.82           & 36.30         & 44.90 \\
&       & \algnameCut & +5.32\upgreen   & +3.16\upgreen   & +8.40\upgreen  & -5.90\downred \\
\bottomrule
\end{tabular}

% \begin{tabular}{l c @{\hspace{4mm}} c c @{\hspace{4mm}} c c}
% \toprule
% \shortstack{\textbf{Model} \\ \textbf{/}\\ \textbf{Data}} & \textbf{LFRQA\%} & \multicolumn{2}{c}{\textbf{Informativeness}} & \multicolumn{2}{c}{\textbf{Truthfulness}} \\
% \cmidrule(r{10pt}){3-4} \cmidrule{5-6}
% & & Bio. & Wild. & Bio. & Wild.\\
% \midrule
% \multirow{4}{*}{\shortstack{Llama3.1\\/\\LFRQA}}
%   & 10\% & 56.40 & 24.90 & 79.79 & 34.50 \\
%   & 40\% & 52.66 & 18.30 & 78.35 & 26.20 \\
%   & 70\% & 50.20 & 17.20 & 75.25 & 21.00 \\
%   & 100\% & 49.82 & 15.80 & 78.43 & 21.00 \\
% \cmidrule(lr){1-6}
% \multirow{4}{*}{\shortstack{Llama3.1\\/\\LFRQA\\+LIMA}}
%    & 0\% & 44.78 & 54.60 & 67.62 & 45.10 \\
%    & 10\% & 51.26 & 29.60 & 75.44 & 34.50 \\
%    & 40\% & 52.05 & 27.20 & 77.02 & 27.50 \\
%    & 70\% & 48.96 & 26.20 & 76.40 & 26.90 \\
%    & 100\% & 45.85 & 24.20 & 75.53 & 27.10 \\
% \midrule[\heavyrulewidth]
% \multirow{4}{*}{\shortstack{Qwen2.5\\/\\LFRQA}}
%    & 10\% & 46.19 & 47.70 & 81.35 & 50.70 \\
%    & 40\% & 46.19 & 40.60 & 80.09 & 44.30 \\
%    & 70\% & 45.22 & 40.50 & 79.39 & 42.50 \\
%    & 100\% & 44.90 & 36.00 & 79.27 & 38.10 \\
% \cmidrule(lr){1-6}
% \multirow{4}{*}{\shortstack{Qwen2.5\\/\\LFRQA\\+LIMA}}
%     & 0\% & 36.07 & 66.50 & 71.14 & 56.00 \\
%     & 10\% & 41.15 & 43.20 & 77.84 & 44.90 \\
%     & 40\% & 49.55 & 41.20 & 78.24 & 41.20 \\
%     & 70\% & 44.06 & 39.20 & 79.52 & 46.40 \\
%     & 100\% & 46.88 & 43.40 & 77.47 & 46.80 \\

% \bottomrule
% \end{tabular}

  }
  \vspace{-2mm}
\caption{Truthfulness (Truth.$\uparrow$) and Informativeness (Info.$\uparrow$) of Llama3.1-8B and Qwen2.5-14B tuned on original (vanilla) or \algnameCut IFT data. For \algnameCut, we report score increase\upgreen\ and decrease\downred\ compared to Vanilla. }
  \label{tab:truth_help_percent}
  \vspace{-1em}
\end{table}

\ifarxiv
\begin{table}[t]
\small
\centering
\begin{tabular}{lcc}
\toprule
 & \textbf{Helpfulness}  &  \textbf{Truthfulness}     \\ \midrule\
Removing Unfamiliar  & 1.57e-3 &  2.91e-11 \\
\bottomrule
\end{tabular}
\vspace{-1mm}
\caption{\label{tab:stat_test_sec2} The p-values of the Wilcoxon Signed-Rank Test on whether Adding LIMA or removing unfamiliar knowledge changes the informativeness and truthfulness of the responses compared to the original LFRQA.}
\vspace{-0.7em}
\end{table}
\else

\fi

Truthfulness and informativeness scores for Vanilla (the original IFT data) and \algnameCut\space (with unfamiliar knowledge removed) IFT are presented in \cref{tab:truth_help_percent}, where we draw the following conclusions:

\myparagraph{Unfamiliar knowledge in high-quality IFT data increases hallucination.} Controlling for all other variables, removing unfamiliar knowledge from IFT data consistently improves truthfulness across all settings. Furthermore, with the amount of LFRQA increases, Llama-3.1-8B Vanilla exhibits a decline in truthfulness for both \emph{Bio} and \emph{WildHalu}, while Qwen2.5-14B Vanilla does not show such a decrease. Therefore, weaker LLMs might be more likely to encounter unfamiliar knowledge in IFT data and are consequently more vulnerable to hallucinations when fine-tuned on such data.

\myparagraph{\algnameCut\space effectively improves truthfulness by removing unfamiliar knowledge.} \algnameCut\space significantly improve truthfulness across all LLMs and IFT data compositions. Notably, the absolute improvement is larger for Llama-3.1-8B than for Qwen2.5-14B, indicating that the algorithm may be particularly beneficial for weaker LLMs that are more vulnerable to hallucination.

\myparagraph{Removing unfamiliar knowledge may reduce informativeness.} Compared to Vanilla IFT, \algnameCut\space reduces the informativeness especially for Llama-3.1-8B. Even for the stronger base model Qwen2.5-14B, informativeness can still decrease, though less frequently.

\myparagraph{Statistical Significance.} Our experiments span a wide range of dataset compositions and LLMs, allowing for robust statistical testing. We conduct Wilcoxon Signed-Rank Tests \citep{Wilcoxon1992}, which indicate removing unfamiliar knowledge using \algnameCut\space significantly reduces informativeness (p-value = 1.66e-3) while improving truthfulness (p-value = 2.33e-11).
%TODO
% \fi

\myparagraph{Takeaway.} With statistical significance, we find that unfamiliar knowledge in high-quality IFT data may cause hallucinations when generalizing to out-of-distribution tasks. \algnameCut\space can effectively improve truthfulness by removing unfamiliar knowledge, but this comes at the potential cost of reduced informativeness---one of the key goals of high-quality IFT.

\section{RQ2: Balancing Informativeness and Truthfulness with \algname} \label{sec:RQ2}

\begin{table*}[t!]
\centering
\resizebox{\textwidth}{!}{

\begin{tabular}{c c c @{\hspace{4mm}} c c c c c @{\hspace{4mm}} c c c c c}
\toprule
\multirow{2.5}{*}{\shortstack{\textbf{Model} \\ \textbf{/}\\ \textbf{Data}}}
  & \multirow{2.5}{*}{\textbf{LFRQA\%}}
  & \multirow{2.5}{*}{\textbf{Method}}
  & \multicolumn{5}{c}{\textbf{Biography}}
  & \multicolumn{5}{c}{\textbf{WildHalu}} \\
\cmidrule(r{13pt}){4-8}\cmidrule(r){9-13}
 & & 
   & Truth.$\uparrow$ 
   & Info.$\uparrow$ 
   & CCP Diff.$\uparrow$ 
   & CCP B.A.$\uparrow$ 
   & Hon. B.A.$\uparrow$
   & Truth.$\uparrow$ 
   & Info.$\uparrow$ 
   & CCP Diff.$\uparrow$ 
   & CCP B.A.$\uparrow$ 
   & Hon. B.A.$\uparrow$ \\
\midrule

\multirow{8}{*}{\shortstack{Llama3.1-8B\\/\\LFRQA}}
 & \multirow{2}{*}{10\%}
   & Vanilla & 56.54 & 22.10 & \grayz & \grayb & \grayb & 82.22 & 29.70 & \grayz & \grayb & \grayb \\
 & 
   & \algname    & -0.14\downred & +2.80\upgreen & 0.1170 & 61.53 & 54.70 & -2.43\downred & +4.80\upgreen & 0.0970 & 60.49 & 52.30 \\
 & \multirow{2}{*}{40\%}
   & Vanilla & 53.41 & 15.60 & \grayz & \grayb & \grayb & 79.00 & 21.90 & \grayz & \grayb & \grayb \\
 & 
   & \algname    & -0.75\downred & +2.70\upgreen & 0.1527 & 63.29 & 53.49 & -0.65\downred & +4.30\upgreen & 0.1182 & 71.66 & 51.34 \\
 & \multirow{2}{*}{70\%}
   & Vanilla & 49.15 & 16.80 & \grayz & \grayb & \grayb & 75.32 & 22.60 & \grayz & \grayb & \grayb \\
 & 
   & \algname    & +1.05\upgreen & +0.40\upgreen & 0.1853 & 70.73 & 54.12 & -0.07\downred & -1.60\downred & 0.1506 & 68.24 & 52.50 \\
 & \multirow{2}{*}{100\%}
   & Vanilla & 50.15 & 15.60 & \grayz & \grayb & \grayb & 73.77 & 22.00 & \grayz & \grayb & \grayb \\
 & 
   & \algname    & -0.33\downred & +0.20\upgreen & 0.1693 & 68.99 & 54.22 & +4.66\upgreen & -1.00\downred & 0.1475 & 71.42 & 51.15 \\
\cmidrule(lr){1-13}

\multirow{8}{*}{\shortstack{Llama3.1-8B\\/\\LFRQA\\+LIMA}}
 % & \multirow{2}{*}{0\%}
 %   & Vanilla & 44.43 & 50.00 & \grayz & \grayb & \grayb & 77.43 & 50.00 & \grayz & \grayb & \grayb \\
 % & 
 %   & \algname    & +0.35\upgreen & +4.60\upgreen & -0.0465 & 50.86 & 51.27 & -9.81\downred & -4.90\downred & 0.0655 & 52.57 & 52.26 \\
 & \multirow{2}{*}{10\%}
   & Vanilla & 50.97 & 34.70 & \grayz & \grayb & \grayb & 79.43 & 32.80 & \grayz & \grayb & \grayb \\
 & 
   & \algname    & +0.29\upgreen & -5.10\downred & 0.1332 & 58.91 & 52.99 & -3.99\downred & +1.70\upgreen & 0.1110 & 63.29 & 51.07 \\
 & \multirow{2}{*}{40\%}
   & Vanilla & 47.51 & 29.90 & \grayz & \grayb & \grayb & 73.89 & 33.50 & \grayz & \grayb & \grayb \\
 & 
   & \algname    & +4.54\upgreen & -2.70\downred & 0.1926 & 66.18 & 54.09 & +3.13\upgreen & -6.00\downred & 0.1568 & 71.62 & 54.07 \\
 & \multirow{2}{*}{70\%}
   & Vanilla & 44.31 & 30.10 & \grayz & \grayb & \grayb & 73.65 & 22.60 & \grayz & \grayb & \grayb \\
 & 
   & \algname    & +4.65\upgreen & -3.90\downred & 0.1470 & 68.99 & 51.81 & +2.75\upgreen & +4.30\upgreen & 0.1316 & 70.92 & 50.47 \\
 & \multirow{2}{*}{100\%}
   & Vanilla & 46.81 & 26.30 & \grayz & \grayb & \grayb & 73.04 & 28.40 & \grayz & \grayb & \grayb \\
 & 
   & \algname    & -0.96\downred & -2.10\downred & 0.1759 & 68.92 & 53.28 & +2.49\upgreen & -1.30\downred & 0.1736 & 70.13 & 51.64 \\
\midrule[\heavyrulewidth]

\multirow{8}{*}{\shortstack{Qwen2.5-14B\\/\\LFRQA}}
 & \multirow{2}{*}{10\%}
   & Vanilla & 41.14 & 46.30 & \grayz & \grayb & \grayb & 79.70 & 42.50 & \grayz & \grayb & \grayb \\
 & 
   & \algname    & +5.05\upgreen & +1.40\upgreen & 0.1805 & 62.44 & 51.89 & +1.65\upgreen & +8.20\upgreen & 0.1334 & 64.73 & 51.73 \\
 & \multirow{2}{*}{40\%}
   & Vanilla & 51.11 & 50.00 & \grayz & \grayb & \grayb & 81.66 & 41.40 & \grayz & \grayb & \grayb \\
 & 
   & \algname    & -4.92\downred & -9.40\downred & 0.1876 & 69.63 & 53.57 & -1.57\downred & +2.90\upgreen & 0.1730 & 70.01 & 51.62 \\
 & \multirow{2}{*}{70\%}
   & Vanilla & 50.64 & 45.30 & \grayz & \grayb & \grayb & 80.86 & 39.80 & \grayz & \grayb & \grayb \\
 & 
   & \algname    & -5.42\downred & -4.80\downred & 0.1888 & 67.17 & 54.63 & -1.47\downred & +2.70\upgreen & 0.1694 & 70.84 & 52.75 \\
 & \multirow{2}{*}{100\%}
   & Vanilla & 49.55 & 47.50 & \grayz & \grayb & \grayb & 80.60 & 39.90 & \grayz & \grayb & \grayb \\
 & 
   & \algname    & -4.65\downred & -11.50\downred & 0.2082 & 71.43 & 55.36 & -1.33\downred & -1.80\downred & 0.1688 & 72.07 & 52.77 \\
\cmidrule(lr){1-13}

\multirow{8}{*}{\shortstack{Qwen2.5-14B\\/\\LFRQA\\+LIMA}}
 % & \multirow{2}{*}{0\%}
 %   & Vanilla & 32.43 & 50.00 & \grayz & \grayb & \grayb & 69.06 & 50.00 & \grayz & \grayb & \grayb \\
 % & 
 %   & \algname    & +3.64\upgreen & +16.50\upgreen & -0.0158 & 52.34 & 48.69 & +2.08\upgreen & +6.00\upgreen & -0.0273 & 51.03 & 51.51 \\
 & \multirow{2}{*}{10\%}
   & Vanilla & 33.48 & 46.60 & \grayz & \grayb & \grayb & 75.05 & 46.20 & \grayz & \grayb & \grayb \\
 & 
   & \algname    & +7.67\upgreen & -3.40\downred & 0.1161 & 56.91 & 50.87 & +2.79\upgreen & -1.30\downred & 0.1276 & 66.64 & 52.90 \\
 & \multirow{2}{*}{40\%}
   & Vanilla & 43.64 & 39.50 & \grayz & \grayb & \grayb & 77.26 & 44.30 & \grayz & \grayb & \grayb \\
 & 
   & \algname    & +5.91\upgreen & +1.70\upgreen & 0.1920 & 63.25 & 53.36 & +0.98\upgreen & -3.10\downred & 0.1959 & 69.10 & 51.87 \\
 & \multirow{2}{*}{70\%}
   & Vanilla & 40.86 & 42.00 & \grayz & \grayb & \grayb & 73.43 & 40.80 & \grayz & \grayb & \grayb \\
 & 
   & \algname    & +3.20\upgreen & -2.80\downred & 0.1972 & 63.60 & 53.57 & +6.09\upgreen & +5.60\upgreen & 0.1897 & 58.86 & 52.21 \\
 & \multirow{2}{*}{100\%}
   & Vanilla & 44.58 & 36.30 & \grayz & \grayb & \grayb & 79.82 & 44.90 & \grayz & \grayb & \grayb \\
 & 
   & \algname    & +2.30\upgreen & +7.10\upgreen & 0.2257 & 68.54 & 54.37 & -2.35\downred & +1.90\upgreen & 0.1922 & 62.99 & 51.84 \\

% \midrule
% \multirow{2}{*}{\shortstack{Qwen2.5-14B\\/\\LFRQA\\+LIMA}} & \algname    & +4.05&  &  &  &  &  &  &  &  & \\

\bottomrule
\end{tabular}

}
\vspace{-2mm}
\caption{\label{tab:section_3} \algname\space vs. Vanilla IFT on two LLMs. Info., Truth., CCP B.A., CCP Diff, and Hon. B.A. denote Informativeness, Truthfulness, CCP Balanced Accuracy, CCP Difference, and Honesty Balanced Accuracy, respectively. We report percentage values of all metrics except CCP Diff. with its actual values. For vanilla IFT, we report random Hon./CCP B.A. and zero CCP Diff. For \algname, we report score increase\upgreen\ and decrease\downred\ compared to Vanilla.} 
\vspace{-0.7em}
\end{table*}

\begin{table}[t]
\small
\centering
\resizebox{\columnwidth}{!}{
\begin{tabular}{lcc}
\toprule
 & \textbf{Informativeness}  &  \textbf{Truthfulness}     \\ \midrule
Using \algname\space Decrease  & 0.4291 & 0.8436 \\
Using \algname\space Increase  & 0.5709 & 0.1607 \\
\bottomrule
\end{tabular}
}
\vspace{-2mm}
\caption{\label{tab:stat_test_sec3} The p-values of the Wilcoxon Signed-Rank Tests on whether using UNIT changes the info. and truth. of responses compared to Vanilla IFT.}
\vspace{-0.7em}
\end{table}

One potential reason for \algnameCut's risk of reducing informativeness is its broad removal of all uncertain claims, which may sometimes strip away valuable content in the original responses that are carefully crafted by prior work.

To strike a balance, we introduce \algname, a variant of \algnameCut\ that preserves the original high-quality responses (e.g., from LIMA and LFRQA) and then adds a \texttt{<reflection>} section to flag the model's uncertain claims. Under \algname, the model is both fine-tuned on the original high-quality responses and on an additional on-policy \texttt{<reflection>} section, thereby learning to reflect on its uncertainty. This approach aims to retain all informative content in IFT data while enhancing truthfulness by marking uncertainty.

\subsection{Methodology - \algname} 
The procedure of \algname\space is illustrated in \cref{fig:illustration}. It shares steps \stepone\ and \steptwo\ with \algnameCut. Their differences lie between steps \stepthree\ and \stepfour\ in how they handle uncertain claims. \algname\ is detailed as follows.

\paragraph{Adding Uncertain Claims to Reflection}
For every instruction-response pair \((I_i,R_i)\) we collect all uncertain claims into an ordered list:
\[
\mathcal{U}_i=\bigl[C_{i,j}\bigr]_{j:\,\ell(C_{i,j})=\text{uncertain}}
\]
We then append uncertain claims $C_{i,j}$ to a \texttt{<reflection>} section following the original response $R_i$, according to the following rules:

\begin{itemize}[leftmargin=1em, label=$\bullet$, noitemsep, partopsep=1pt, topsep=2pt, parsep=1pt]%[leftmargin=*,label=$\bullet$,noitemsep,partopsep=0pt,topsep=0pt,parsep=0pt]
  \item \textbf{No uncertain claims} (\(|\mathcal{U}_i|=0\)):  
        append  
        \emph{``I am confident about the accuracy and the truthfulness of the information provided.''}
  \item \textbf{Moderate uncertainty} (\(1\le |\mathcal{U}_i|\le T\)):  
        list the \(|\mathcal{U}_i|\) uncertain claims in bullet form so users can see which points the model is uncertain about.
  \item \textbf{High uncertainty} (\(|\mathcal{U}_i|>T\)):  
        append  
        \emph{``I am unconfident about the accuracy and the truthfulness of most of the information provided above.''}
\end{itemize}

% \vspace{3pt}
The verbosity threshold $T$ limits the number of uncertain claims shown in the \texttt{<reflection>} section, ensuring the output remains concise and easy to interpret. In our experiments, we set it to 10. Templates of how we construct \texttt{<reflection>} are available in \cref{app:unit_ref_templates}.

\subsection{Evaluation Metrics} \label{sec:3_metrics}
\noindent \textbf{Truthfulness and Informativeness.} Since \algname\ does not modify the content of the original IFT responses but appends a \texttt{<reflection>} section, we expect that---when the \texttt{<reflection>} section is excluded---its truthfulness and informativeness will remain comparable to those of vanilla IFT. To test this hypothesis, we evaluate the truthfulness and informativeness of only the answer component from models tuned by \algname, with the \texttt{<reflection>} section removed, using the same metrics as in \cref{sec:rq1_ex_details}.

\myparagraph{CCP Balanced Accuracy.} \algname\space aims to teach the model to recognize and explicitly label uncertainty. We assess whether uncertain claims are correctly placed in the \texttt{<reflection>} while certain claims are left unreflected. In other words, CCP B.A. measures how "well" the model learns its own uncertainty boundary. We define \emph{CCP Balanced Accuracy} as:
% \vspace{-5pt}
\[
\text{CCP B.A.} 
= \frac{1}{2}\Biggl( 
    \frac{\lvert UC_{\text{reflected}} \rvert}{\lvert UC_{\text{all}} \rvert} 
    + \frac{\lvert CC_{\text{unreflected}} \rvert}{\lvert CC_{\text{all}} \rvert} 
\Biggr)
% \vspace{-3pt}
\]
where \(\lvert UC_{\text{reflected}} \rvert\) is the number of reflected uncertain claims, \( \lvert UC_{\text{all}} \rvert\) is the total number of uncertain claims, \(\lvert CC_{\text{unreflected}} \rvert \) is the number of unreflected certain claims, and \(\lvert CC_{\text{all}} \rvert\) is the total number of certain claims. Here, ``uncertain'' and ``certain'' are determined by the CCP threshold (75th percentile) used during training.

\myparagraph{CCP Difference.}
Besides learning to classify uncertain claims by a threshold, the model could learn to rank claims by their CCP scores. To assess this behavior, we compute the difference in the mean CCP of reflected claims versus that of unreflected claims. A positive CCP Difference indicates that the model reflects more often on more uncertain claims than certain claims, and vice versa.

\myparagraph{Honesty Balanced Accuracy.}
To evaluate how reliably the model reflects factually incorrect claims while leaving correct claims unreflected, we compute \emph{Honesty Balanced Accuracy}, which follows the same formula as CCP Balanced Accuracy but uses claim correctness as gold labels instead of CCP-based uncertainty. 

A more detailed description of the evaluation metrics is available at  \cref{app:eval_metric_details}.

\subsection{Experiment Results}

We compare \algname\space with vanilla IFT in all data combinations in \cref{sec:RQ1}. Results are presented in \cref{tab:section_3}. Our key observations are:

\myparagraph{\algname\space maintains informativeness and truthfulness compared to vanilla IFT.} 
Cohering to its algorithmic design, \algname\space does not significantly compromise the informativeness or truthfulness of the answer part of the response (without reflection) compared to vanilla IFT. 
We conduct the Wilcoxon Signed-Rank Test \citep{Wilcoxon1992} to confirm the statistical significance of UNIT's influence on the informativeness and truthfulness of the response. As shown, \cref{tab:stat_test_sec3} indicates no statistically significant differences in both informativness and truthfulness of the response compared to vanilla IFT.

\myparagraph{Models tuned with \algname\ recognise uncertainty, leading to better honesty.} We observe a positive CCP Difference, and CCP Balanced Accuracy significantly above random (50\%). This suggests that the models can learn and predict their claim-level uncertainty to some extent. 
Furthermore, \algname\space achieves above-random Honesty Balanced Accuracy. This indicates that uncertainty reflections help mitigate hallucinations by warning users about uncertain claims, thereby informing them about the likelihood and location of potential hallucinations.
Compared to CCP, Honesty B.A. shows a smaller gain over the random baseline, likely because uncertainty does not always indicate factual correctness \citep{fadeeva-etal-2024-fact}, we discuss this in detail in \cref{limitation}.

\begin{table}[t]
  \centering
  \setlength{\tabcolsep}{3pt} % tighten column padding
  \resizebox{\columnwidth}{!}{%
    \begin{tabular}{l c c c @{\hspace{4mm}} c c}
  \toprule
  \multirow{2.5}{*}{\shortstack{\textbf{Model} \\ \textbf{/}\\ \textbf{Data}}}
    & \multirow{2.5}{*}{\textbf{LFRQA\%}}
    & \multicolumn{2}{c}{\textbf{Biography}}
    & \multicolumn{2}{c}{\textbf{WildHalu}} \\
  \cmidrule(r{6pt}){3-4}\cmidrule(r){5-6}
   & 
    & Hon. B.A.$\uparrow$ & Upper Bound
    & Hon. B.A.$\uparrow$ & Upper Bound \\
  \midrule
  \multirow{4}{*}{\shortstack{Llama3.1-8B\\/\\LFRQA}}
    & 10\%  & 54.70 & 62.94 & 52.30 & 60.42 \\
    & 40\%  & 53.49 & 64.13 & 51.34 & 61.06 \\
    & 70\%  & 54.12 & 65.72 & 52.50 & 61.45 \\
    & 100\% & 54.22 & 64.45 & 51.15 & 61.37 \\
  \midrule
  \multirow{5}{*}{\shortstack{Llama3.1-8B\\/\\LFRQA\\+LIMA}}

    & 0\%   & 51.27 & 61.72 & 52.26 & 61.67 \\
    & 10\%  & 52.99 & 62.07 & 51.07 & 62.45 \\
    & 40\%  & 54.09 & 62.39 & 54.87 & 62.55 \\
    & 70\%  & 51.81 & 62.47 & 50.47 & 63.53 \\
    & 100\% & 53.28 & 62.97 & 51.64 & 63.01 \\
  \midrule
  \multirow{4}{*}{\shortstack{Qwen2.5-14B\\/\\LFRQA}}
    & 10\%  & 51.89 & 60.67 & 51.73 & 57.06 \\
    & 40\%  & 53.57 & 66.11 & 51.62 & 58.39 \\
    & 70\%  & 54.63 & 64.17 & 52.75 & 60.89 \\
    & 100\% & 55.36 & 66.05 & 52.77 & 62.26 \\
  \midrule
  \multirow{5}{*}{\shortstack{Qwen2.5-14B\\/\\LFRQA\\+LIMA}}

    & 0\%   & 48.69 & 61.51 & 51.51 & 60.75 \\
    & 10\%  & 50.87 & 58.35 & 52.90 & 60.31 \\
    & 40\%  & 53.36 & 64.85 & 51.87 & 59.10 \\
    & 70\%  & 53.57 & 64.68 & 52.21 & 60.24 \\
    & 100\% & 54.37 & 64.57 & 51.84 & 59.49 \\
  \bottomrule
\end{tabular}
  }
  \vspace{-2mm}
  \caption{Comparison of Honesty Balanced Accuracy (Hon.\,B.A.) of \algname\ and its upper‐bound performance for Biography and WildHalu of Llama3.1-8B fine-tuned under various settings.}
  \label{tab:honba_upper}
   \vspace{-0.7em}
\end{table}

\subsection{Upper Bound of \algname}
%Even with \algname-tuning, Honesty Balanced Accuracy is only slightly above the random baseline (50\%). 
The performance of \algname\space on Honesty Balanced Accuracy can be influenced by several factors, for example (1) uncertainty-factuality mismatch: intuitively, LLM uncertainty relates to, but does not always indicate, factual accuracy. An uncertain guess to a question might be correct while a confident claim might also be wrong. (2) Imperfect uncertainty measurement, as shown by \citet{vashurin2025benchmarkinguncertaintyquantificationmethods}, uncertainty quantification is a very challenging task. Biased uncertainty scores may have less predictive power for factual inaccuracy. 
To find the highest possible Honesty Balanced Accuracy using CCP, we calculate the test-time CCP of all claims and search for the best CCP threshold. Results are demonstrated in \cref{tab:honba_upper}. The upper bounds of Honesty Balanced Accuracy rarely exceed 65, showing that achieving high Honesty Balanced Accuracy is difficult even with the ground truth CCP ranking and the best thresholding. Therefore, given the achievable honesty upper bound, \algname\space performs reasonably well in Honesty B.A. Future improvements in uncertainty measurement \citep{vashurin2025benchmarkinguncertaintyquantificationmethods} may further enhance its performance.

\section{Conclusion}
In this paper, we investigate how unfamiliar knowledge in Instruction Fine-tuning (IFT) affects LLM truthfulness. We first propose \algnameCut, to remove unfamiliar knowledge and fine-tune the LLM on an unfamiliar knowledge-free IFT dataset. \algnameCut\ substantially improves the truthfulness of LLM but at the cost of risking informativeness. To strike a better balance, we introduce \algname, which preserves the original responses and appends a ``reflection'' section that flags uncertain claims. Empirically, \algname\ maintains both informativeness and truthfulness compared to vanilla IFT. Unlike prior work in honesty alignment that relies on constructing honesty-specific training data, our methods demonstrate that LLM honesty can be improved directly using existing high-quality IFT datasets, which addresses the critical need of practitioners who rely on IFT to adapt LLMs to downstream tasks and domains. Our work highlights another interesting use case for uncertainty measurements---besides quantifying the uncertainty of LLM generations, they can also be leveraged to measure LLMs' familiarity to existing high-quality data, which might be essential for fine-tuning.

\section*{Limitations}\label{limitation}
\myparagraph{Uncertainty's Limited Indication on Truthfulness.}
In this work, we improve the honesty of LLMs by teaching them about their own uncertainty, following the definition of honesty from the previous work \citep{Park2023AIDA, yang2024alignmenthonesty}. However, uncertainty does not perfectly indicate factual accuracy and thus may hinder the performance of algorithms predicting uncertainty signals. A model may be honest—faithfully reflecting its beliefs (even if those beliefs are incorrect), resulting in untruthful output. This distinction also sheds light on the results in \cref{tab:section_3} where the improvement on CCP metrics are higher than the honesty balanced accuracy. To address this limitation, future work may investigate the differences between uncertain and factually wrong or leverage uncertainty estimations that can better indicate task-specific factuality \citep{vashurin2025benchmarkinguncertaintyquantificationmethods}. 

\myparagraph{Limitation in Existing Uncertainty Measurements.}
In this study, we use CCP to measure claim-level uncertainty. CCP is the state-of-the-art uncertainty measurement that shows the best effectiveness on information-seeking tasks \citep{fadeeva-etal-2024-fact}. However, it also shows limitations on other tasks, as discussed in \cref{sec:unit_cut} and \citet{vashurin2025benchmarkinguncertaintyquantificationmethods}. Furthermore, CCP scores yield relative measures of uncertainty but do not provide a deterministic threshold to distinguish “uncertain” from “certain” claims, for which we employed the 75th percentile of training-data CCP scores as a heuristic cutoff. These factors are essential for apply \algname\ and \algnameCut\ to a broader scale, but are out of the scope of this paper. We leave these explorations to future work.

\ifarxiv
\section*{Acknowledgements} 
This paper has received funding from the Swiss
National Science Foundation (SNSF) under the project `How sustainable is sustainable finance? Impact evaluation and automated greenwashing detection' (Grant Agreement No. 100018\_207800). It is also funded by grant from Hasler Stiftung for the Research Program Responsible AI with the project ``Scientific Claim Verification.'' 
\else

\fi
\section*{Ethics Statement}
\myparagraph{Data Privacy or Bias.} We use publically available IFT datasets which have no data privacy issues or bias against certain demographics. All artifacts we use are under licenses allowing research usage. We also notice no ethical risks associated with this work.
\myparagraph{Reproducibility Statement.} To ensure full reproducibility, we will disclose all codes and data used in this project, as well as the LLM generations. For OpenAI models, using \texttt{gpt-4o-2024-11-20} and \texttt{gpt-4o-mini-2024-07-18} with random seed 42 will ensure reproducing the observations in paper, but not the exact numbers due to the poor reproducibility of OpenAI API.
\bibliography{custom}

\appendix

\section{Uncertainty for Different Types of Instructions} \label{app:why_info_only}
We showcase a toy example in \cref{sec:unit_cut} explaining why CCP or any other claim-level uncertainty measurement are proposed for information-seeking tasks only. Limited by the current SoTA of uncertainty measurement, it is impossible to extend \algname\space or \algnameCut\space to non-information-seeking tasks. Here we give two other examples from LIMA where CCP might be ill-defined. Moreover, there are roughly 25\% of LIMA data in reasoning, math, or coding, making it hard for CCP to measure factual uncertainty.

More than 20\% of datapoints in LIMA are creative writing, where the CCP uncertainty clearly does not indicate factual uncertainty or knowledge familiarity. Example 1 below shows one of them:
\begin{lstlisting}[frame=single, basicstyle=\ttfamily\scriptsize, xleftmargin=0pt, numbers=none]
Prompt: Write a story of a perfectly ordinary or boring day except write it as dramatically as possible.

Response: Posted before, but here's my average morning from last year:
1:30am. Something has woken me up from my slumber. It screams, it howls, it calls my name. Stumbling from my bed, I stagger down the hallway to the source of the noise. The monster wants food. I placate it with a drink and scurry back to my warm, cosy bed, hoping it will last until the sun comes up.
6:50am. Nothing. No noise. The monster is sleeping. It will need waking before long, to capture it and strap it into it's travelling cage.
7:30am. The monster is secured. We begin our journey to deposit the alternate care-giver to his location. Once he is delivered, we have to press on, continuing through the hunger pangs.
8:05am. The hunger pangs are consuming us. We stop to refuel our bellies. The monster objects. It is temporarily released into the wild, but soon recaptured.
8:30am. We have arrived at the monsters location for the day. It is left with new care-givers, a list of instructions handed out. It is fed, and I leave.
8:55am. Freedom.
9:00am. A day of filing paperwork, away from a screeching toddler. Bliss.....
\end{lstlisting}

Even for information-seeking tasks, CCP sometimes faces challenges. Example 2 below shows a response where many niche information are provided in a table, which may induce hallucination. However, CCP's design cannot check uncertainty for information in tables.
\begin{lstlisting}[frame=single, basicstyle=\ttfamily\scriptsize, xleftmargin=0pt, numbers=none]
Prompt: View tabular file such as CSV from command line, having horizontal and vertical scrolling would be great.

Response: Sure, please take a look at csvkit. It provides a set of tools that adhere to the UNIX philosophy (meaning they are small, simple, single-purposed and can be combined).

Here is an example that extracts the ten most populated cities in Germany from the free Maxmind World Cities database and displays the result in a console-readable format:
```$ csvgrep -e iso-8859-1 -c 1 -m "de" worldcitiespop | csvgrep -c 5 -r "\d+"
  | csvsort -r -c 5 -l | csvcut -c 1,2,4,6 | head -n 11 | csvlook
-----------------------------------------------------
|  line_number | Country | AccentCity | Population  |
-----------------------------------------------------
|  1           | de      | Berlin     | 3398362     |
|  2           | de      | Hamburg    | 1733846     |
|  3           | de      | Munich     | 1246133     |
|  4           | de      | Cologne    | 968823      |
|  5           | de      | Frankfurt  | 648034      |
|  6           | de      | Dortmund   | 594255      |
|  7           | de      | Stuttgart  | 591688      |
|  8           | de      | Dusseldorf | 577139      |
|  9           | de      | Essen      | 576914      |
|  10          | de      | Bremen     | 546429      |
-----------------------------------------------------
```

Csvkit is platform independent because it is written in Python.
\end{lstlisting}

\section{Prompts for Training Data Construction} \label{app:info_seeking_prompt}

\subsection{Atomic Claim Extraction}
{\small
\begin{tcolorbox}[
    breakable,
    colframe=boxborder,    % Border color
    colback=boxbg,         % Background color
    coltitle=boxtitle,     % Title text color
    colbacktitle=boxheader,% Title background color
    fonttitle=\bfseries,   % Bold title font
    % sharp corners,         % No rounded edges
    boxrule=1pt,           % Border thickness
    width=\columnwidth,  % Fit within half-column
    title=Atomic Claim Generation Prompt, % Box title
    enhanced,              % Enable more styles
    arc=5pt,               % Slightly rounded edges
    left=6pt, right=6pt, top=4pt, bottom=4pt % Padding
]

Break down the following sentence into atomic facts.\\
\_\_\_\\
{sentence}\\
\_\_\_\\
\\
Respond with the following format:\\
\\
- <atomic fact 1>\\
- <atomic fact 2>\\
...\\
\\
However, if there is no factual claim, respond <EMPTY>.
\end{tcolorbox}
}

\subsection{Classifying Instruction Type}
We take the instruction classification prompt from \citet{xu2024magpiealignmentdatasynthesis}, which is illustrated below. We deemed the instruction to be "information-seeking" if only if the "primary\_tag" is "Information seeking" and "other\_tags" is empty.

{\small

\begin{tcolorbox}[
    breakable,
    colframe=boxborder,
    colback=boxbg,
    coltitle=boxtitle,
    colbacktitle=boxheader,
    fonttitle=\bfseries,
    boxrule=1pt,
    width=\columnwidth,  % Full width
    title=Info-Seeking Classification Prompt Template,
    enhanced,
    arc=5pt,
    left=6pt, right=6pt, top=4pt, bottom=4pt
]

\# Instruction\\
Please label the task tags for the user query.\\
\#\# User Query\\
\{USER QUERY\}\\
\#\# Tagging the user input\\
Please label the task tags for the user query. You will need to analyze the user query and select the most relevant task tag from the list below.\\
all\_task\_tags = [\\
"Information seeking", \# Users ask for specific information or facts about various topics.\\
"Reasoning", \# Queries require logical thinking, problem\-solving, or processing of complex ideas.\\
"Planning", \# Users need assistance in creating plans or strategies for activities and projects.\\
"Editing", \# Involves editing, rephrasing, proofreading, or other tasks related to the composition of general written content.\\
"Coding \& Debugging", \# Users seek help with writing, reviewing, or fixing code in programming.\\
"Math", \# Queries related to mathematical concepts, problems, and calculations.\\
"Role playing", \# Users engage in scenarios requiring ChatGPT to adopt a character or persona.\\
"Data analysis", \# Requests involve interpreting data, statistics, or performing analytical tasks.\\
"Creative writing", \# Users seek assistance with crafting stories, poems, or other creative texts.\\
"Advice seeking", \# Users ask for recommendations or guidance on various personal or professional issues.\\
"Brainstorming", \# Involves generating ideas, creative thinking, or exploring possibilities.\\
"Others", \# Any queries that do not fit into the above categories or are of a miscellaneous nature.\\
]\\
\#\# Output Format:\\
Note that you can only select a single primary tag. Other applicable tags can be added to the list of other tags.\\
Now, please output your tags below in a json format by filling in the placeholders in <...>:\\
\\
\{\{\\
"primary\_tag": "<primary tag>",\\
"other\_tags": ["<tag 1>", "<tag 2>", ... ]\\
\}\}
\end{tcolorbox}
}

\subsection{Rewriting Certain Claims to New Reponses}
In \algnameCut\space, we use \texttt{gpt-4o-2024-11-20} to rewrite a list of atomic claims into new responses. The prompt is presented below:

\section{Evaluation Metrics Details}
\label{app:eval_metric_details}

\myparagraph{Truthfulness Score.} We use the database and information retriever of FactScore \citep{min-etal-2023-factscore} and WildFactScore\citep{zhao2024wildhallucinationsevaluatinglongformfactuality} to conduct retrieval-augmented fact-checking. We follow \citet{min-etal-2023-factscore} but replace \texttt{gpt-3.5-turbo} with \texttt{gpt-4o-mini} for the evaluation model. The prompts for generating atomic claims and fact-checking are listed below.

{\small
\begin{tcolorbox}[
    colframe=boxborder,    % Border color
    colback=boxbg,         % Background color
    coltitle=boxtitle,     % Title text color
    colbacktitle=boxheader,% Title background color
    fonttitle=\bfseries,   % Bold title font
    % sharp corners,         % No rounded edges
    boxrule=1pt,           % Border thickness
    width=\columnwidth,  % Fit within half-column
    title=Fact-Checking Prompt, % Box title
    enhanced,              % Enable more styles
    arc=5pt,               % Slightly rounded edges
    left=6pt, right=6pt, top=4pt, bottom=4pt % Padding
]

Analyze the following question and its associated claim:\\
\\
Question: \{input\}\\
\\
Claim: \{claim\}\\
\\
Some context that might be helpful to fact-check the Claim:\\
\{context\}\\
\\
Now answer: is all information provided in the <claim> true given the context and your latest knowledge?\\
\end{tcolorbox}
}

\citet{min-etal-2023-factscore} use heuristics to decide if there is ``True'' or ``False'' in LLMs' fact-checking response, while we leverage the following prompt to summarize fact-checking outcome, which should be more accurate.

{\small
\begin{tcolorbox}[
    colframe=boxborder,    % Border color
    colback=boxbg,         % Background color
    coltitle=boxtitle,     % Title text color
    colbacktitle=boxheader,% Title background color
    fonttitle=\bfseries,   % Bold title font
    % sharp corners,         % No rounded edges
    boxrule=1pt,           % Border thickness
    width=\columnwidth,  % Fit within half-column
    title=Fact-Checking Summarization Prompt, % Box title
    enhanced,              % Enable more styles
    arc=5pt,               % Slightly rounded edges
    left=6pt, right=6pt, top=4pt, bottom=4pt % Padding
]

Question: \{input\}\\
\\
Claim: \{claim\}\\
\\
Is the above claim true?\\
\\
Reply: \{reply\}\\
\\
Summarize this reply into one word, whether the claim is true: "True", "False" or "Not known".
\end{tcolorbox}
}

% \vspace{1em}
\myparagraph{Informativeness Score.} We adapt the prompt from MT-Bench \citep{zheng2023judgingllmasajudgemtbenchchatbot} for informativeness evaluation, which is shown as below. To mitigate LLM-judge position bias, we compute informativeness scores for both original and swapped pairs of (target answer, reference answer). For tie-breaking, if one judgement says ``A/B wins'' and another says ``Tie'', the final judge is ``A/B wins'' as one judge leans towards A or B. If one judgement says ``A/B wins'' but another says ``B/A wins'' reversely, the final judge is ``Tie'' as there is no clear tendency.

{\small
\begin{tcolorbox}[
    colframe=boxborder,    % Border color
    colback=boxbg,         % Background color
    coltitle=boxtitle,     % Title text color
    colbacktitle=boxheader,% Title background color
    fonttitle=\bfseries,   % Bold title font
    % sharp corners,         % No rounded edges
    boxrule=1pt,           % Border thickness
    width=\columnwidth,  % Fit within half-column
    title=Helpfulness Judging Prompt, % Box title
    enhanced,              % Enable more styles
    arc=5pt,               % Slightly rounded edges
    left=6pt, right=6pt, top=4pt, bottom=4pt % Padding
]

Please act as an impartial judge and evaluate the quality of the responses provided by two AI assistants to the user question displayed below. You should choose the assistant that follows the user's instructions and answers the user's question better. Your evaluation should focus on factors such as the helpfulness, relevance, depth, and level of detail of their responses. Do not take correctness into consideration. Begin your evaluation by comparing the two responses and provide a short explanation. Avoid any position biases and ensure that the order in which the responses were presented does not influence your decision. Do not allow the length of the responses to influence your evaluation. Do not favor certain names of the assistants. Be as objective as possible. After providing your explanation, output your final verdict by strictly following this format: "[[A]]" if assistant A is better, "[[B]]" "if assistant B is better, and "[[C]]" for a tie.\\
\\
\#\#\# User's Question:\\
\{question\}\\
\\
<|The Start of Assistant A's Response to the User|>\\
\{answer\_a\}\\
<|The End of Assistant A's Response to the User|>\\
<|The Start of Assistant B's Response to the User|>\\
\{answer\_b\}\\
<|The End of Assistant B's Response to the User|>\\

\end{tcolorbox}
}
\myparagraph{CCP Balanced Accuracy.}
We evaluate LLMs' ability to model uncertainty by calculating the CCP Balanced Accuracy. First, using the Atomic Claims Generation Prompt template from \cref{app:eval_metric_details}, we extract all answer claims from the model's response, denoted as \(AC_{\text{all}}\). Next, we employ GPT-as-a-judge with the prompt template shown below to identify the atomic claims reflected in the response's \texttt{<reflection>} section, denoted as \(AC_{\text{reflected}}\). 
{\small
    \begin{tcolorbox}[
        breakable,
        colframe=boxborder,
        colback=boxbg,
        coltitle=boxtitle,
        colbacktitle=boxheader,
        fonttitle=\bfseries,
        boxrule=1pt,
        width=\columnwidth,  % Full width
        title=Get $AC_{reflected}$ Prompt Template,
        enhanced,
        arc=5pt,
        left=6pt, right=6pt, top=4pt, bottom=4pt
    ]

\#\#\# Instruction\\
You will be given a question and two list relating to the question, claim list and reflection list that was extracted from an answer to the question.\\
Please help to extract two new list from the claim list and the reflection list:\\
1. Covered Claims: All the claims in Claim list that is COVERED by at least one of the reflections in reflection list.\\
2. Covered Reflection: All the reflections in reflection list that is COVERED by at least one of the claims in Claim list.\\

For Example:\\
- Question:\\
Tell me a bio of Cheyenne Brando.\\

- Claim List:\\
Cheyenne Brando was born in 1996.\\
Cheyenne Brando is the daughter of Marlon Brando.\\
Cheyenne Brando is the daughter of Tarita Teriipaia.\\
She was born in Tahiti.\\
Her parents lived in Tahiti after they married.\\
Her parents married following the filming of Mutiny on the Bounty.\\
She has a half-sister named Miko.\\
Miko is from Brando's relationship with his second wife.\\
Brando's second wife is Movita Castaneda.\\
Cheyenne Brando is named after a character.\\
Cheyenne Brando's father has a character in The Wild One.\\

- Reflection List:\\
Marlon Brando was an actor.\\
Marlon Brando had a relationship with Movita Castaneda.\\
Miko is a half-sister of Cheyenne Brando.\\
Cheyenne Brando is named after her father's character in The Wild One.\\

\# Output\\
- Covered Claims:\\
She has a half-sister named Miko.\\
Brando's second wife is Movita Castaneda.\\
Cheyenne Brando is named after a character.\\
Cheyenne Brando's father has a character in The Wild One.\\

- Covered Reflection:\\
Marlon Brando had a relationship with Movita Castaneda.\\
Miko is a half-sister of Cheyenne Brando.\\
Cheyenne Brando is named after her father's character in The Wild One.\\

Now it's your turn to answer, follow the format in the example strictly:\\
- Question:\\
\{USER'S INSTRUCTION\}\\

- Claim List:\\
\{$AC_{reflected}$\}\\

- Reflection List:\\
\{ClAIMS FROM <reflection>\}\\

\end{tcolorbox}
}

Then, by applying the CCP method with the 75th quantile threshold from the training data, we label the uncertain answer claims, denoted as \(UC_{\text{all}}\). From these sets, we derive:
\begin{itemize}[label={}, leftmargin=0pt, itemsep=0pt, topsep=0pt]
    \item \textbf{CCP TP (Reflected Uncertain Claims):}\\ \( UC_{\text{reflected}} = AC_{\text{reflected}} \cap UC_{\text{all}} \)
    \item \textbf{CCP TN (Unreflected Certain Claims):}\\ \( CC_{\text{unreflected}} = (AC_{\text{all}} \setminus AC_{\text{reflected}}) \setminus UC_{\text{all}} \)
    \item \textbf{CCP TN+FP (Certain Claims):}\\ \( CC_{\text{all}} = AC_{\text{all}} \setminus UC_{\text{all}} \)
    \item \textbf{CCP TP+FN (Unertain Claims):} \( UC_{\text{all}}\)
\end{itemize}
% \vspace{5pt}
CCP Balanced Accuracy is then computed as:
\[
\text{CCP B.A.} = \frac{1}{2}\left( \frac{\lvert UC_{\text{reflected}} \rvert}{\lvert UC_{\text{all}} \rvert} + \frac{\lvert CC_{\text{unreflected}} \rvert}{\lvert CC_{\text{all}} \rvert} \right)
\]

\myparagraph{Honesty Balanced Accuracy.}
Honesty Balanced Accuracy is computed similarly to CCP Balanced Accuracy, but instead of using uncertainty labels, we use truthfulness labels obtained from FactScore and WildFactScore (see \cref{app:eval_metric_details}). First, each atomic claim in the response is labeled as \emph{True} or \emph{False} based on its factual correctness. Let:
\begin{itemize}[label={}, leftmargin=0pt, itemsep=0pt, topsep=0pt]
    \item \( TC_{\text{all}} \) be the set of all true claims.
    \item \( FC_{\text{all}} \) be the set of all false claims.
\end{itemize}
Next, we identify the true claims that were reflected in the response:
\[
TC_{\text{reflected}} = AC_{\text{reflected}} \cap TC_{\text{all}}
\]
and the false claims that were not reflected in the response:
\[
FC_{\text{unreflected}} = (AC_{\text{all}} \setminus AC_{\text{reflected}}) \cap FC_{\text{all}}
\]
Honesty Balanced Accuracy is then defined as:
\[
\text{Honesty B.A.} = \frac{1}{2}\left( \frac{\lvert TC_{\text{reflected}} \rvert}{\lvert TC_{\text{all}} \rvert} + \frac{\lvert FC_{\text{unreflected}} \rvert}{\lvert FC_{\text{all}} \rvert} \right)
\]

\myparagraph{CCP Difference.}
CCP difference measures the model's ability to learn the ranking claims with their uncertainty (CCP scores). This is computed by the difference between the average CCP of the reflected answer claims \(AC_{reflected}\) and the average CCP of the unreflected answer claims \(AC_{reflected}\). A positive CCP Difference indicates that the reflected claims are more uncertain compared to the unreflected claims on average, and vice versa.

% This is an appendix.
% 
\section{Experiment Implementation Details} \label{app:experiment_details}

\subsection{Hyperparameter Settings}
For experiments in this paper, we conducted full fine-tuning on Llama-3.1-8B \cite{grattafiori2024llama3herdmodels} for 3 epochs with 2 NVIDIA H100-80GB. We utilized "The Alignment Handbook" code base released by Huggingface to fine-tune all the models \cite{Tunstall_The_Alignment_Handbook}. The configurations of our hyper-parameters are detailed in Table \ref{tab:UNIT hyperparameter}.

\begin{table}[h]
    \centering
    \small % Reduce font size for better fit
    \renewcommand{\arraystretch}{1.2} % Adjust row spacing
    \setlength{\tabcolsep}{3pt} % Reduce column spacing
    \begin{tabular}{ll}
        \toprule
        \textbf{Configuration} & \textbf{UNIT} \\
        \midrule
        Model & Llama-3.1-8B \\
        Number of epochs & 3 \\
        Devices & 2 H100 GPU (80 GB) \\
        Total Batch size & 32 samples \\
        Optimizer &Paged AdamW 32bit \\ & \cite{Loshchilov2017FixingWD} \\
        Scheduler & Cosine\\
        Learning rate & $1 \times 10^{-5}$ \\
        Warmup Ratio & 0.03 \\
        \bottomrule
    \end{tabular}
    \caption{Training Configuration for UNIT}
    \label{tab:UNIT hyperparameter}

\end{table}

We used the default chat template in "The Alignment Handbook" \cite{Tunstall_The_Alignment_Handbook} for fine-tuning all models, as illustrated below.

{\small
\begin{tcolorbox}[
    colframe=black,    
    colback=boxbg,         
    coltitle=boxtitle,     
    colbacktitle=gray,
    fonttitle=\bfseries,
    boxrule=1pt,
    width=0.48\textwidth,
    title=Fine-tuning Chat Template,
    enhanced,
    arc=5pt,
    left=6pt, right=6pt, top=4pt, bottom=4pt
]
<|system|>\\
\{SYSTEM\_PROMPT\}
<|end\_of\_text|>\\
<|user|>\\
\{USER\_PROMPT\}
<|end\_of\_text|>\\
<|assistant|>\\
\{ASSISTANT\_RESPONSE\}
<|end\_of\_text|>
\end{tcolorbox}
}

\subsection{Inference}
For our LLM inference tasks, we employ vLLM \cite{kwon2023efficient} with the following configuration: a temperature setting of 0, a repetition penalty of 1, and a maximum output of 2048 tokens.

\subsection{Information-seeking Data Filtering}

\subsection{System Prompts}

In fine-tuning, we used different system prompts for surgery and non-surgery data points. For surgery data points, we used the following system prompt:

{\small
\begin{tcolorbox}[
    colframe=black,    % Border color
    colback=boxbg,         % Background color
    coltitle=boxtitle,     % Title text color
    colbacktitle=gray,% Title background color
    fonttitle=\bfseries,   % Bold title font
    % sharp corners,         % No rounded edges
    boxrule=1pt,           % Border thickness
    width=0.48\textwidth,  % Fit within half-column
    title=System Prompt for Surgery Data Points, % Box title
    enhanced,              % Enable more styles
    arc=5pt,               % Slightly rounded edges
    left=6pt, right=6pt, top=4pt, bottom=4pt % Padding
]

You are a helpful assistant.\\
you should answer user's query first, providing a helpful and accurate response. Then write a <reflection> section following your response, listing all the factual claims you made in your response that you are uncertain about.\\
\\
Output your reflection in the following format ONLY:\\
<reflection>\\
The following summarizes the facts that I am uncertain about in my answer:\\
1. [factual claim 1 that you are uncertain about]\\
2. [factual claim 2 that you are uncertain about]\\
3. [factual claim 3 that you are uncertain about]\\
...[more factual claims]...\\
\end{tcolorbox}
}

For non-surgery data points, we used the following system prompt:

{\small
\begin{tcolorbox}[
    colframe=black,    % Border color
    colback=boxbg,         % Background color
    coltitle=boxtitle,     % Title text color
    colbacktitle=gray,% Title background color
    fonttitle=\bfseries,   % Bold title font
    % sharp corners,         % No rounded edges
    boxrule=1pt,           % Border thickness
    width=0.48\textwidth,  % Fit within half-column
    title=System Prompt for Non-Surgery Data Points, % Box title
    enhanced,              % Enable more styles
    arc=5pt,               % Slightly rounded edges
    left=6pt, right=6pt, top=4pt, bottom=4pt % Padding
]

You are a helpful assistant.\\
you should answer user's query directly, providing a helpful and accurate response to the query.
\end{tcolorbox}
}

\section{Details of Templates used in \algname} \label{app:unit_ref_templates}

% \begin{figure}[h!]
{\small
\begin{tcolorbox}[
    colframe=boxborder,    % Border color
    colback=boxbg,         % Background color
    coltitle=boxtitle,     % Title text color
    colbacktitle=boxheader,% Title background color
    fonttitle=\bfseries,   % Bold title font
    % sharp corners,         % No rounded edges
    boxrule=1pt,           % Border thickness
    width=0.48\textwidth,  % Fit within half-column
    title=Surgery Template 1, % Box title
    enhanced,              % Enable more styles
    arc=5pt,               % Slightly rounded edges
    left=6pt, right=6pt, top=4pt, bottom=4pt % Padding
]

\{ORIGINAL RESPONSE\}\\
\\
<reflection>:\\
The following summarizes the facts that I am uncertain about in my answer:\\
1. \{UNCERTAIN CLAIM 1\}\\
2. \{UNCERTAIN CLAIM 2\}\\
...

\end{tcolorbox}
}
% \vspace{-6pt}
% \caption{}
% \label{fig:surgery_template1}
% \vspace{-1em}
% \end{figure}

% \begin{figure}[h!]
{\small
\begin{tcolorbox}[
    colframe=boxborder,    % Border color
    colback=boxbg,         % Background color
    coltitle=boxtitle,     % Title text color
    colbacktitle=boxheader,% Title background color
    fonttitle=\bfseries,   % Bold title font
    % sharp corners,         % No rounded edges
    boxrule=1pt,           % Border thickness
    width=0.48\textwidth,  % Fit within half-column
    title=Surgery Template 2, % Box title
    enhanced,              % Enable more styles
    arc=5pt,               % Slightly rounded edges
    left=6pt, right=6pt, top=4pt, bottom=4pt % Padding
]

\{ORIGINAL RESPONSE\}\\
\\
<reflection>:\\
I am unconfident about the accuracy and the truthfulness of most of the information provided above.

\end{tcolorbox}
}
% \vspace{-6pt}
% \caption{}
% \label{fig:surgery_template2}
% \vspace{-1em}
% \end{figure}

% \begin{figure}[h!]
{\small
\begin{tcolorbox}[
    colframe=boxborder,    % Border color
    colback=boxbg,         % Background color
    coltitle=boxtitle,     % Title text color
    colbacktitle=boxheader,% Title background color
    fonttitle=\bfseries,   % Bold title font
    % sharp corners,         % No rounded edges
    boxrule=1pt,           % Border thickness
    width=0.48\textwidth,  % Fit within half-column
    title=Surgery Template 3, % Box title
    enhanced,              % Enable more styles
    arc=5pt,               % Slightly rounded edges
    left=6pt, right=6pt, top=4pt, bottom=4pt % Padding
]

\{ORIGINAL RESPONSE\}\\
\\
<reflection>:\\
I am confident about the accuracy and the truthfulness of the information provided.

\end{tcolorbox}
}
% \caption{}
% \label{fig:surgery_template3}
% \vspace{-1em}
% \end{figure}

\section{LFRQA$_{certain}$ and LFRQA+LIMA$_{certain}$ Construction} \label{app:certain_construction}
In this section, we detail the construction of LFRQA$_{certain}$ and LFRQA+LIMA$_{certain}$ in detail.

To construct LFRQA$_{certain}$ and LFRQA+LIMA$_{certain}$, we use the same approach in \algname\space to find the uncertain claims in each response. To keep the readability after removing all the uncertain claims, we used GPT-4o to remove all the uncertain claims within the original 
response. The prompt template we used is provided as shown below. %in Figure \ref{fig:remove_uncertain_prompt}.

% \begin{figure}[h!]
{\small
\begin{tcolorbox}[
    colframe=boxborder,    % Border color
    colback=boxbg,         % Background color
    coltitle=boxtitle,     % Title text color
    colbacktitle=boxheader,% Title background color
    fonttitle=\bfseries,   % Bold title font
    % sharp corners,         % No rounded edges
    boxrule=1pt,           % Border thickness
    width=0.48\textwidth,  % Fit within half-column
    title=Prompt Template for Removing Uncertain Claims, % Box title
    enhanced,              % Enable more styles
    arc=5pt,               % Slightly rounded edges
    left=6pt, right=6pt, top=4pt, bottom=4pt % Padding
]

[Instruction]: "\{INSTRUCTION\}"\\

[Fact List]: """\{FACT LIST\}"""\\

Please concatenate the facts from the [Fact List] to form a helpful [Response] to the [Instruction].\\

Important Requirements:\\
1. Make sure your [Response] sounds helpful, fluent, and natural. Use logical conjunctions frequently.\\
2. Do not add new fact or information except from those in [Fact List].\\
3. Make sure to involve all information in [Fact List].\\

[Response]:

\end{tcolorbox}
}

\begin{table}[h]
    \centering
    \small
    \begin{tabular}{ccc}
        \toprule
        \textbf{Quantile} & \textbf{LIMA} & \textbf{LFRQA} \\
        \midrule
        0.50 & -0.217175  & -0.052052 \\
        0.65 & -0.086788  & -0.011424 \\
        0.75 & -0.037325  & -0.002476 \\
        0.85 & -0.008926  & -0.000260 \\
        0.95 & -0.000382  & -0.000005 \\
        \bottomrule
    \end{tabular}
    \caption{Comparison of CCP Values at Different Quantiles between LIMA and LFRQA (info-seeking only)}
    \label{tab:data_ccp_detail}
\end{table}

\begin{table}[h]
    \centering
    \small
    \begin{tabular}{lcc}
        \toprule
         & \textbf{LIMA} & \textbf{LFRQA} \\
        \midrule
        \textbf{\# Data Points}                      & 1022         & 14016 \\
        \textbf{\# Info-Seeking Data Point}          & 171  & 14016 \\
        \textbf{Avg. \# of claims per Data Points}   & 44.35  & 8.558\\
        \textbf{Avg. Response Length}               & 435.83  & 79.47 \\
        \bottomrule
    \end{tabular}
    \caption{Data Details of LIMA and LFRQA}
    \label{tab:data_general_detail}
\end{table}

The details of the two datasets are shown in \cref{tab:data_ccp_detail} and \cref{tab:data_general_detail}.

\begin{table}[h]
\small
\centering
\begin{tabular}{lcccc}

\toprule
& \multicolumn{2}{c}{Avg. Truth\,$|\Delta|$} 
& \multicolumn{2}{c}{Avg. Info\,$|\Delta|$} \\
\cmidrule(lr){2-3}\cmidrule(lr){4-5}
& Bio & Wild & Bio & Wild \\
\midrule
\algnameCut & 11.32 & 6.38 & 8.78 & 8.79 \\
\algname & 3.49  & 3.14 & 5.19 & 3.96 \\
\bottomrule
\end{tabular}
\caption{\label{tab:avg_abs_change} Comparison of the average absolute changes of \algnameCut\ and \algname\ relative to vanilla IFT in Truthfulness and Informativeness.}
% \vspace{-1.0em}
\end{table}

\end{document}